\documentclass[sigconf,natbib=false]{acmart}
\AtBeginDocument{%
  }

\setcopyright{acmlicensed}

\copyrightyear{2026}
\acmYear{2026}
\acmDOI{XXXXXXX.XXXXXXX}
\acmConference[SAC'26]{The 41st ACM/SIGAPP Symposium on Applied Computing}{March 23--27, 2026}{Thessaloniki, Greece}
\acmISBN{979-X-XXXX-XXXX-X/26/03}

\RequirePackage[
  datamodel=acmdatamodel,
  style=acmnumeric,
  ]{biblatex}

\begin{document}

\title{Improving LLM-based Ontology Matching with fine-tuning on synthetic data}
\renewcommand{\shorttitle}{LLM-based OM with fine-tuning}

 \author{Guilherme H. Santos Sousa}
 \orcid{0000-0002-2896-2362}
 \affiliation{%
   \institution{IRIT \& Universit\'e de Toulouse 2 Jean Jaur\`es}
   \city{Toulouse} 
   \country{France}
   \postcode{31400}  
 }
 \email{Guilherme.Santos-Sousa@irit.fr}

 \author{Rinaldo Lima}
 \affiliation{%
   \institution{Universidade Federal Rural de Recife}
   \city{Recife} 
   \state{Pernambuco} 
   \country{Brazil}
   \postcode{ 52171-900 }  
 }
 \email{rjl4@cin.ufpe.br}

 \author{Cassia Trojahn}
 \affiliation{%
   \institution{Univ. Grenoble Alpes, Inria, CNRS, Grenoble INP}
   \postcode{38000}
   \city{Grenoble} 
   \country{France}}
 \email{cassia.trojahn-dos-santos@univ-grenoble-alpes.fr}

\renewcommand{\shortauthors}{Sousa et al.}

\begin{abstract}
Large Language Models (LLMs) are increasingly being integrated into various components of Ontology Matching pipelines. This paper investigates the capability of LLMs to perform ontology matching directly on ontology modules and generate the corresponding alignments. Furthermore, it is explored how a dedicated fine-tuning strategy can enhance the model's matching performance in a zero-shot setting. The proposed method incorporates a search space reduction technique to select relevant subsets from both source and target ontologies, which are then used to automatically construct prompts. Recognizing the scarcity of reference alignments for training, a novel LLM-based approach is introduced for generating a synthetic dataset. This process creates a corpus of ontology submodule pairs and their corresponding reference alignments, specifically designed to fine-tune an LLM for the ontology matching task. The proposed approach was evaluated on the Conference, Geolink, Enslaved, Taxon, and Hydrography datasets from the OAEI complex track. The results demonstrate that the LLM fine-tuned on the synthetically generated data exhibits superior performance compared to the non-fine-tuned base model. The key contribution is a strategy that combines automatic dataset generation with fine-tuning to effectively adapt LLMs for ontology matching tasks.
\end{abstract}



\maketitle
\section{Introduction}

Knowledge Graphs (KGs) are powerful structures for representing relational data, modeling how entities are interconnected within a specific domain. A key challenge in integrating and reusing these data structures arises from schema heterogeneity, where two KGs covering the same topic are modeled with different underlying schemas. These differences prevent direct integration and create the need for an alignment between the graphs. Heterogeneities can be linguistic (using different languages or synonyms for the same concept) or structural: representing the same idea with different levels of detail or composition. For example, a common structural heterogeneity is when the concept \texttt{AcceptedPaper} in one KG is equivalent to the combination of \texttt{Paper} and \texttt{Acceptance} in another. Without a formal alignment to bridge such gaps, seamlessly merging the two KGs is impossible.

Traditional matching methods, which rely on lexical comparisons or naive semantic similarity measures, are often insufficient to address the complex heterogeneities between ontologies. A significant advancement was achieved through the integration of embeddings—dense vector representations generated by Language Models or graph encoding techniques. These embeddings provide a more nuanced semantic similarity by capturing the contextual meaning of entities, effectively resolving issues like homonymy that purely lexical metrics cannot handle. Despite their power and practicality, relying solely on embedding similarity is inadequate for correctly matching complex entities, particularly those that map to a specific composition of multiple entities in the target ontology. A canonical example of this challenge is the correspondence between a \texttt{FullName} entity in a source ontology and the pair of \texttt{FirstName} and \texttt{LastName} in a target ontology. While embeddings can identify a high semantic similarity between FullName and the individual FirstName and LastName concepts, they are incapable of defining the structural transformation required to form an equivalence. In this scenario, the semantics of FullName can only be replicated by concatenating FirstName and LastName in a specific order. This compositional rule is rarely explicit in the ontology, requiring the matching system to infer and formalize this transformation during the matching process.

Large Language Models (LLMs) now dominate research in Knowledge Graph (KG) and ontology matching. Their ability to directly generate alignments from ontologies provided in a prompt offers remarkable flexibility. This allows matching systems to generalize across diverse domains and enables developers to instruct the model to perform various semantic tasks within the matching pipeline simply by modifying the prompt. Consequently, it is now feasible to create adaptable matching frameworks where different LLMs can be interchanged, allowing for upgrades to more powerful or domain-specific models. While the application of LLMs for simple one-to-one entity matching is becoming widespread, their exploration for discovering complex alignments remains an emerging area of research.

However, directly applying LLMs to complex matching introduces significant challenges. First, complex entities are effectively subgraphs of their respective KGs, causing the search space for potential alignments to grow exponentially and requiring effective search space reduction strategies. Second, providing the entire source and target ontologies as input can be computationally prohibitive, consuming vast resources depending on the KG's size. Finally, a critical requirement is the ability to produce alignments in a standardized format, such as the Expressive and Declarative Ontology Alignment Language (EDOAL) \cite{DBLP:journals/semweb/DavidESS11}, to facilitate automatic evaluation. Many LLMs lack familiarity with such formalisms, as their training data may not include examples of the EDOAL syntax, hindering their ability to generate verifiable and machine-readable output.

One of the initial works to explore LLMs in complex matching is the paper by \cite{DBLP:conf/kgswc/AminiNHA24}, in which the GMO ontology was provided as context to ChatGPT to find corresponding modules in the GBO ontology. This pioneering study highlighted critical challenges: the prohibitive resource requirements for processing entire ontologies with locally hosted LLMs, and the generation of alignments in unstructured natural language, which requires a manual evaluation process. Building on this, the work by \cite{DBLP:conf/om2/SousaLT24} addressed some of these limitations by proposing a search space reduction strategy coupled with few-shot prompting. This method successfully guided the model to produce alignments in the structured EDOAL format while reducing resource consumption. However, while effective in generating correct EDOAL syntax and identifying simple alignments, its performance on complex alignments still needs improvement.

In this work, it is proposed to enhance LLM performance on complex matching by applying instruction fine-tuning. As demonstrated in \cite{DBLP:conf/iclr/WeiBZGYLDDL22}, fine-tuning can lead to superior performance and better task adaptation without relying on in-context examples. The proposed approach is rigorously evaluated on multiple datasets from the OAEI complex track.

The remainder of this paper is organized as follows. Section \ref{sec:related} provides an overview of related methods for complex matching and discusses their limitations. Section \ref{sec:Approach} introduces the proposed approach, which integrates fine-tuning with LLMs to enhance the understanding and performance of complex alignment tasks. Section \ref{settings} contains the experiment settings. Then, a series of experiments and case studies are presented in Section \ref{results}, highlighting both the strengths and weaknesses of this approach. Finally, Section \ref{conclusion} offers concluding remarks and outlines directions for future research, stressing the potential of fine-tuned LLMs in advancing the state of the art in complex semantic matching.

\section{Related Work}
\label{sec:related}
Recent advancements in ontology matching can be broadly categorized into two main streams of research: approaches based on entity and graph embeddings, and those ones based on LLMs.

\subsection{Embedding-based}

Embedding-based matchers primarily use vector representations to compute semantic similarity between ontology subgraphs. For example, the work in \cite{DBLP:journals/corr/abs-2502-13619} extends the CANARD matcher \cite{DBLP:journals/semweb/ThieblinSHT24} by incorporating LLM-generated embeddings to improve the matching of entities retrieved within the context of SPARQL queries. This enhanced method aggregates four main types of embeddings: label similarity, SPARQL query representations, subgraph embeddings, and instance embeddings. The objective of this aggregation is to address the challenges of representing complex entities and thereby achieve more accurate results. However, the CANARD architecture's reliance on fixed triple and path structures restricts its pattern-matching capabilities, as it cannot identify correspondences that require differently structured subgraphs.

Another matcher incorporating embeddings \cite{DBLP:conf/ecai/SilvaFP24} proposes a novel approach to complex multi-ontology matching (CMOM), presenting a holistic matching solution for complex cases. This method combines lexical string similarity with geometric operations on a shared semantic space, derived from LLM embeddings, to discover complex mappings that involve multiple entities from different ontologies. The process involves several steps: preprocessing ontology vocabularies, generating candidate mappings through both lexical and LLM-based methods, and finally, aggregating and filtering these candidates to obtain the final alignment. A key limitation, however, is that this approach identifies corresponding target entities for a source entity without specifying how they logically combine to reconstruct its semantics. This results in a less general solution, as different logical constructors are often required to formally express the meaning of complex correspondences.

\subsection{LLM-based}

The emergence of LLMs has established a dominant paradigm in ontology matching research \cite{DBLP:conf/kcap/HertlingP23,DBLP:conf/esws/GiglouDEA24}. 
A notable example of this trend is OntoAligner \cite{DBLP:conf/esws/GiglouDKA25}, a Python toolkit designed to integrate traditional methods with contemporary AI techniques, including Retrieval-Augmented Generation (RAG) and the direct application of LLMs. This toolkit provides a flexible and extensible framework for ontology matching, featuring a modular architecture that allows users to customize alignment algorithms, incorporate new datasets, and fine-tune pipelines for diverse use cases. However, a significant limitation is that the framework is primarily designed for simple alignments, focusing on the discovery of one-to-one (1:1) correspondences between source and target entities.

One of the earliest applications of LLMs to the complex matching task was presented in \cite{DBLP:conf/kgswc/AminiNHA24}, who applied ChatGPT-4 to match the GMO and GBO ontologies from the Geolink dataset. Their work was among the first to propose loading entire ontologies directly into the prompt and instructing the LLM to generate the complete alignment. However, this strategy has significant limitations that hinder its general applicability. First, loading complete ontologies is computationally expensive and often infeasible due to the context length and memory constraints of modern LLMs. Second, the model was prompted to generate alignments in unstructured natural language, which precludes direct automatic evaluation. These factors severely compromise the approach's scalability and evaluability, as the lack of a standardized output format needs manual post-processing. These issues were subsequently addressed in \cite{DBLP:conf/om2/SousaLT24}, who introduced a novel approach combining a search space reduction strategy with the generation of final alignments in a structured format. Their method tackles the challenge of processing large ontologies by selecting and integrating only the most relevant subsets from the source and target ontologies into the prompt. This reduction is achieved by automatically generating SPARQL queries based on entity PageRank scores. These queries, in conjunction with an embedding strategy, are used to retrieve semantically similar entities and their local graph structures. The resulting subsets are then presented to the LLM, which is prompted to generate complex alignments directly in the structured EDOAL format, rather than using natural language descriptions. This direct, standardized output simplifies the automatic evaluation and verification of the resulting alignments. While the approach is effective at reducing prompt size and enforcing a structured response, its performance in generating high-quality complex alignments still needs improvement.

\section{Approach}
\label{sec:Approach}
\begin{figure}
    \centering
    \includegraphics[width=0.65\linewidth]{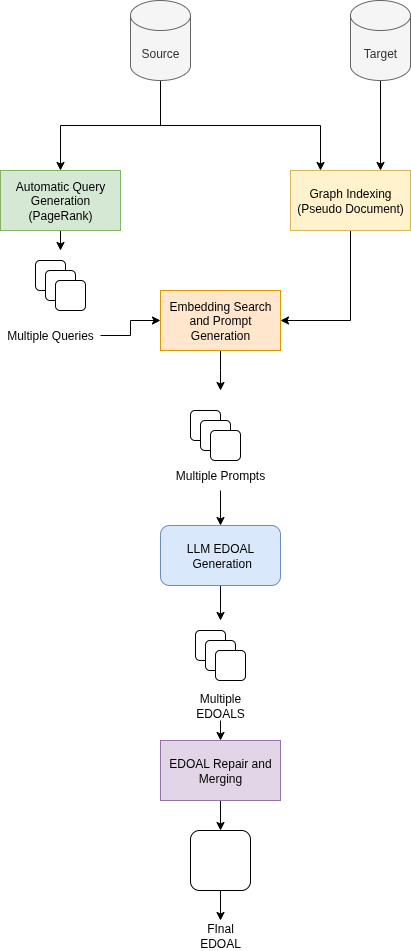}
    \caption{Overview of the proposed LLM-based ontology alignment pipeline. Starting from a source and target ontology, queries are automatically generated from the source ontology using a PageRank-based method. These queries are then used to select subontology parts to construct multiple prompts for direct LLM alignment, which generate multiple EDOAL alignments. These alignments are then post-processed through a repair and merging step to produce the final EDOAL alignment.
}
    \label{fig:enter-label}
\end{figure}

This work extends a previous method proposed in the literature \cite{DBLP:conf/om2/SousaLT24} by adding an instruction fine-tuning phase. Synthetically generated data is used to boost the performance of LLMs on the ontology matching task. The core idea is to make the problem manageable by breaking it down. The proposed method follows a clear pipeline: (i) \textbf{Decompose}: first, a space reduction strategy splits large ontologies into smaller, focused subontologies;(ii) \textbf{Query}: next,  an LLM is prompt with these subontologies to generate partial alignments; and (iii) \textbf{Merge}: finally, the individual outputs are merged to create the final alignment. To measure the impact of these contributions, this strategy is applied and tested in both zero-shot and fine-tuned settings. The overall architecture is illustrated in Figure \ref{fig:enter-label} and the proposed method for generating the synthetic data is described in the subsections below.

\subsection{Space Reduction and Prompt Construction}
The ontology alignment task requires comparing a vast search space of potential entity pairs, a process that is not only computationally expensive but also infeasible to handle within the context window of a single LLM prompt. To overcome this limitation, the proposed method employs a space reduction strategy. This strategy leverages both structural and semantic features to partition the ontologies into smaller, more relevant modules (i.e., subontologies), making the matching task tractable for an LLM.

The process unfolds in several stages. First, structurally significant entities are identified within the source ontology using centrality metrics such as PageRank. For each of these core source entities, embedding-based similarity is used to retrieve a candidate set of semantically related entities from the target ontology. Next, local modules are extracted around these corresponding source and target entity sets by including their neighboring classes and properties. Each of these paired modules forms a reduced subontology pair, which is then inserted into a predefined prompt template. This template instructs the LLM to find equivalent entities and return the partial alignment in EDOAL format. By iterating this process, multiple partial alignments are generated covering the most salient parts of the ontologies. These results are subsequently aggregated and post-processed to construct the final, comprehensive alignment for the entire ontology pair.

\subsection{Alignment Aggregation}

In this approach, the LLM is not just one component in a larger pipeline; it performs the entire matching process after the initial space reduction. In this process, each LLM call processes a pair of subontologies and generates a partial alignment in EDOAL. These partial outputs are then merged into a single alignment file after generation. To handle redundancy and potential conflicts across prompts, a post-processing step is included to normalize, deduplicate, and validate the alignment. This step ensures that repeated mappings across different subontology pairs are not counted multiple times and that the final output conforms to the required format for evaluation.

\subsection{Dataset Generation}

\begin{figure}
    \centering
    \includegraphics[width=\linewidth]{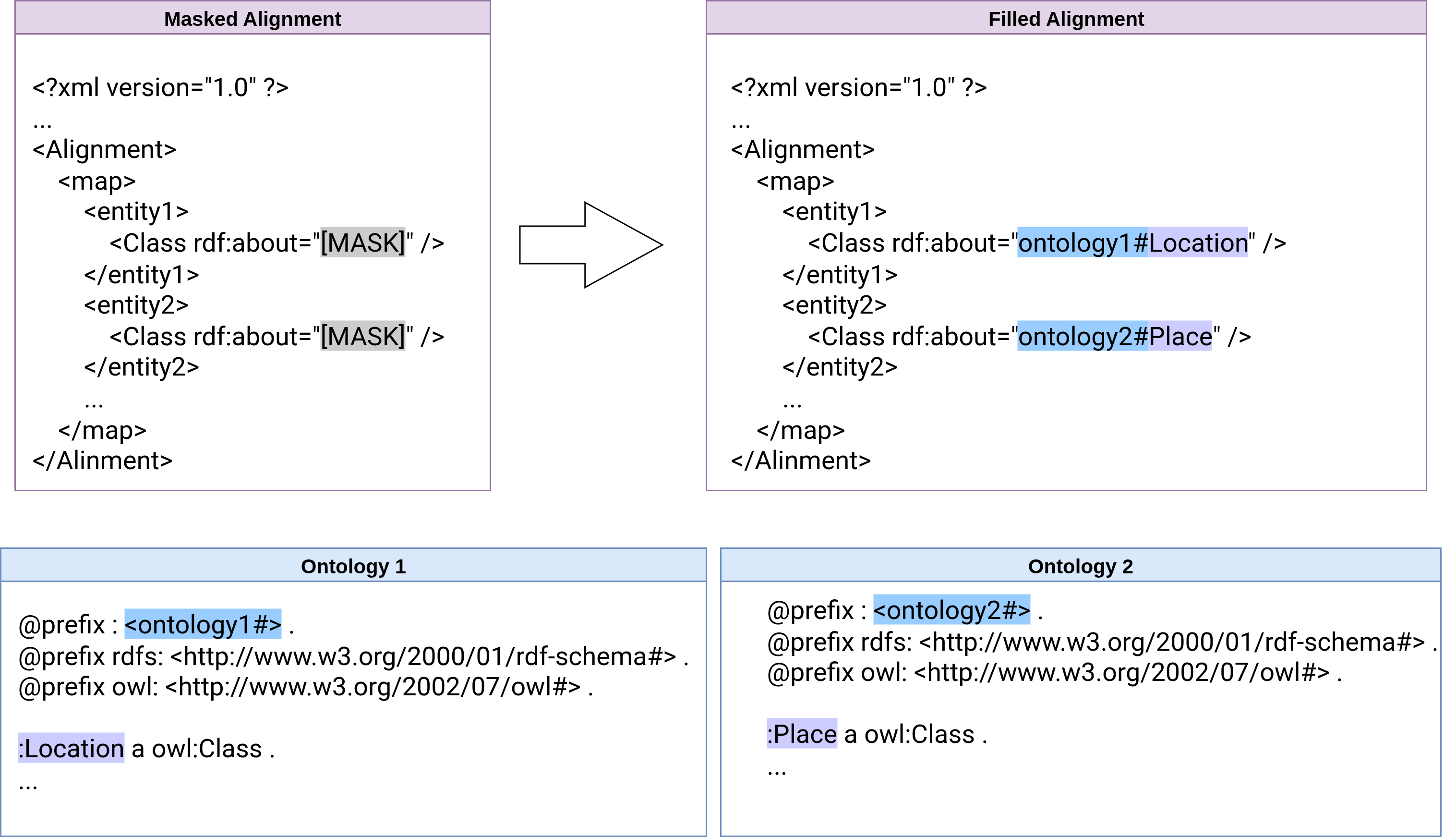}
    \caption{Alignment and ontology generation workflow. Starting from a seed alignment template, the LLM is prompted to fill in the masked placeholders with entities while preserving semantic equivalence. Once the alignment is completed, the LLM is then prompted to sequentially generate the ontologies that correspond to the proposed alignment.}
    \label{fig:fill}
\end{figure}

To improve the LLM's understanding of the ontology matching task, an instruction fine-tuning strategy is applied using an LLM-generated training dataset. The synthetic data generation method is based on the hypothesis that predicting the ontologies given the alignment between them is easier than finding the alignment between two ontologies, which is the objective task to solve. Since most of the models do not know the EDOAL structure, giving a seed EDOAL structure without the entities mitigates this problem, as the LLM just needs to fill in the blanks with plausible entity names in the EDOAL structure (not generate complex syntax from scratch), as illustrated in figure \ref{fig:fill}. Once the alignment template is filled, the LLM is prompted to generate the ontologies where the previous alignment was generated, trying to include entities not in common between the two ontologies. To provide contrast, a separate prompt instructs the LLM to generate ontology pairs without entities in common. This is a very important step for teaching the models to learn when they should not create an alignment, which helps them learn to avoid false positives. When building the dataset, there is no guarantee that the filled entities are semantically coherent and that equivalence holds between the generated entities. However, the hypothesis is that if fine-tuning on this noisy dataset still leads to improved performance on real-world benchmarks, it proves the model is learning valuable patterns for the ontology matching task.

The seed EDOAL structure is generated procedurally by deriving sentences \cite{jm3} from the grammar present on the EDOAL webpage by randomly selecting a production rule from the grammar to build the template.\footnote{http://ns.inria.org/edoal/1.0/} The code for this generator, along with all evaluations, is available on \url{https://anonymous.4open.science/r/llm-6E88/}. In addition to the synthetic data, the approach was evaluated using cross-validation with a dataset of manually created modules. For each fold, the model is fine-tuned on a set of subontology pairs and their corresponding ground-truth EDOAL alignments. This training data includes a mix of both simple and complex mappings to ensure broad coverage. The LLM's core task is to learn how to generate the correct EDOAL mappings directly from any given pair of ontology fragments.

\section{Experiment Settings}
\label{settings}
To evaluate the fine-tuning approach and compare it with the baselines, a series of experiments using pretrained LLMs was conducted. This section details the datasets, models, experimental configurations, and metrics used in this evaluation. The evaluation was performed on five datasets from the OAEI 2020 Complex Matching track:\footnote{https://oaei.ontologymatching.org/2020/results/complex/index.html}: \textbf{Conference}, \textbf{Geolink}, \textbf{Enslaved}, \textbf{Taxon}, and \textbf{Hydrography}. This specific year was chosen because it features the largest number of participating datasets and matching systems, providing a larger comparison. These datasets contain a variety of subjects, making them ideal for testing the limits of both baseline and fine-tuned models.

Distinct LLMs were employed for different stages of this experiment. For the Dataset Generation step, Microsoft/Phi-4\footnote{https://huggingface.co/microsoft/phi-4} was used. For the baseline experiment (b0), the models are Meta/LLaMA-3.2-3B-Instruct (3B), Meta/LLaMA-3.1-8B-Instruct (8B), microsoft/Phi-4-mini-instruct (4B), microsoft/phi-4 (14B), Qwen/Qwen3-14B (14B), a reasoning model, and mistralai/Mistral-7B-Instruct-v0.3 (7B). All the models were downloaded from HuggingFace\footnote{https://huggingface.co/models} and run locally. For the fine-tuning experiments, the performance of two setups was compared using the model the models Meta/LLaMA-3.2-3B-Instruct (3B) due to resource constraints:

\textbf{Cross-Validation Fine-Tuning (b1)}: In this "leave-one-out" approach, the model is fine-tuned on data from four of the datasets and then evaluated on the remaining one;
\textbf{Synthetic Data Fine-Tuning (b2)}: In this setup, the model is fine-tuned on the synthetically generated dataset.

For all variants, the LLM was prompted using a multi-turn format. The system role established the task, and the user provided the module pair for matching. The modules were generated manually with the procedure described in the next section to compare only the LLM performance without the effect of the space reduction module. Performance was measured using the metrics proposed in \cite{DBLP:conf/esws/SousaLS25}. These metrics adapt the standard precision, recall, and F-measure to effectively evaluate complex alignments while also applying to simple (1:1) correspondences. The generated alignments from each model were compared against the gold standard reference alignments to compute the final scores.

\section{Results and Discussion}
\label{results}

To improve the capacity of the matcher in producing complex alignments, the point of higher increase in performance is in the LLM module in the architecture. To verify the impact of the model on the architecture performance, it needs to be evaluated in isolation from the other modules, as the reduction of space can remove or insert irrelevant entities that can change the results. To this, a set of modules was manually created for all 5 ontologies, and then, for all entities in all correspondence, the modules were merged to form the input ontology. The number of modules for all ontologies is present in Table \ref{tab:ontology-modules}. Furthermore, all correspondences with overlapping modules are joined in a single alignment. With this procedure is ensured that all entities in the correspondences are included in the input ontologies. 

\begin{table}[ht]
\centering
\begin{tabular}{|l |r|}
\hline
\textbf{Ontology} & \textbf{Modules} \\
\hline
cmt & 12 \\
ekaw & 15 \\
edas & 24 \\
confOf & 13 \\
conference & 22 \\
enslaved & 18 \\
wikidata & 19 \\
gbo & 24 \\
gmo & 31 \\
cree & 13 \\
swo & 22 \\
hydrOntology & 35 \\
hydro3 & 7 \\
taxon & 4 \\
agrovoc & 2 \\
taxref & 4 \\
dbpedia & 2 \\
\hline
\end{tabular}
\caption{Ontology modules count}
\label{tab:ontology-modules}
\end{table}

\begin{figure}
    \centering
    \includegraphics[width=\linewidth]{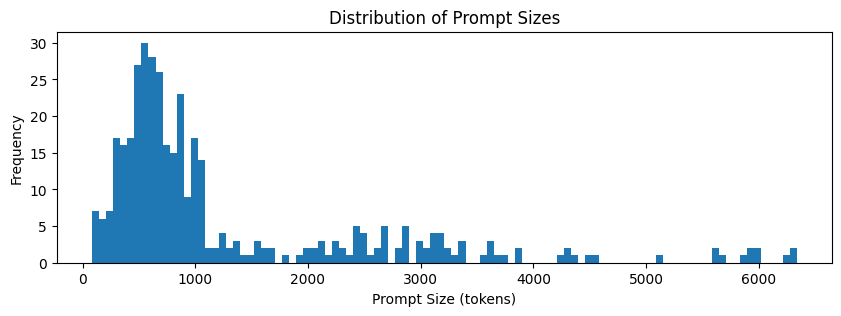}
    \caption{Distribution of prompt tokens .}
    \label{fig:nmod-dist}
\end{figure}

The procedure of creation of the modules is the following: for all entities in the module, if the entity has no rdf:type and is from a standard vocabulary from http://www.w3.org or http://ns.inria.org/edoal/ it is filtered. Then the entity is added to the new ontology, with descriptions and labels added. For simplicity, properties with BNode as objects are filtered, and 5 superclasses are added, and then the final modules are rendered in turtle. This procedure was used to keep the maximum prompt size within a manageable range. With this approach, the maximum prompt size is 6336 tokens from the hydrography dataset in the pair hydrOntology-swo, and most of the prompts lie in the range between 0 and 1000 tokens. Estimating the prompt sizes by just concatenating the two ontologies, arrives in the distribution present in figure \ref{fig:nmod-dist}. With those modules, it is possible to experiment with fine-tuning in the LLM, isolating it from the impact on the automatic module generation performance.

A preliminary experiment was done with the LLMs Meta/LLaMA-3.1-8B-Instruct (8B) and microsoft/phi-4 (14B) to see how the models behave if given the manually produced modules. This experiment was conducted by using the baseline approach and another prompt template that uses complex correspondence templates described in \cite{scharffe2009correspondence}. The results of this preliminary evaluation are presented in Table \ref{tab:llama_phi_full_results}. It is possible to see in the results that both approaches have better results in the simple case when compared to the complex case. Considering the LLMs, the phi-4 that has more parameters performed better on average in the simple case, while having comparable results in the complex case. The correspondence patterns in the prompt result in slightly lower performance in both simple and complex cases when compared with the base approach.

\begin{table}[ht]
\centering
\scriptsize
\begin{tabular}{|l|l|cccccc|}
\hline
\textbf{Type} & \textbf{Dataset} & \textbf{A prec.} & \textbf{B prec.} & \textbf{A rec.} & \textbf{B rec.} & \textbf{A f1} & \textbf{B f1} \\
\hline
Simple  & Conference   & 0.25 & 0.42 & 0.54 & 0.58 & 0.28 & 0.43 \\
                 & Enslaved     & 0.16 & 0.14 & 0.20 & 0.18 & 0.16 & 0.14 \\
                 & Geolink      & 0.20 & 0.22 & 0.19 & 0.18 & 0.17 & 0.18 \\
                 & Hydrography  & 0.21 & 0.24 & 0.33 & 0.37 & 0.16 & 0.21 \\
                 & Taxon        & 0.06 & 0.15 & 0.08 & 0.00 & 0.03 & 0.00 \\
\hline
Complex  & Conference  & 0.01 & 0.04 & 0.11 & 0.08 & 0.00 & 0.01 \\
                  & Enslaved    & 0.01 & 0.00 & 0.06 & 0.08 & 0.01 & 0.00 \\
                  & Geolink     & 0.05 & 0.03 & 0.17 & 0.14 & 0.05 & 0.02 \\
                  & Hydrography & 0.00 & 0.01 & 0.21 & 0.21 & 0.00 & 0.01 \\
                  & Taxon       & 0.02 & 0.00 & 0.18 & 0.14 & 0.03 & 0.00 \\
\hline
Simple  & Conference   & 0.15 & 0.36 & 0.39 & 0.50 & 0.18 & 0.38 \\
(Patterns)                 & Enslaved     & 0.07 & 0.16 & 0.12 & 0.15 & 0.08 & 0.14 \\
                 & Geolink      & 0.13 & 0.20 & 0.16 & 0.19 & 0.12 & 0.15 \\
                 & Hydrography  & 0.11 & 0.22 & 0.20 & 0.31 & 0.09 & 0.19 \\
                 & Taxon        & 0.13 & 0.25 & 0.13 & 0.10 & 0.09 & 0.09 \\
\hline
Complex   & Conference   & 0.06 & 0.02 & 0.07 & 0.06 & 0.02 & 0.01 \\
(Patterns)                  & Enslaved     & 0.01 & 0.00 & 0.05 & 0.09 & 0.01 & 0.00 \\
                  & Geolink      & 0.03 & 0.01 & 0.12 & 0.11 & 0.03 & 0.01 \\
                  & Hydrography  & 0.05 & 0.01 & 0.11 & 0.15 & 0.02 & 0.00 \\
                  & Taxon        & 0.00 & 0.01 & 0.07 & 0.16 & 0.01 & 0.01 \\
\hline
\end{tabular}
\caption{Comparison of A (Llama 3.1) and B (Phi 4) on Simple and Complex datasets with and without patterns.}
\label{tab:llama_phi_full_results}
\end{table}

\begin{figure}
    \centering
    \includegraphics[width=\linewidth]{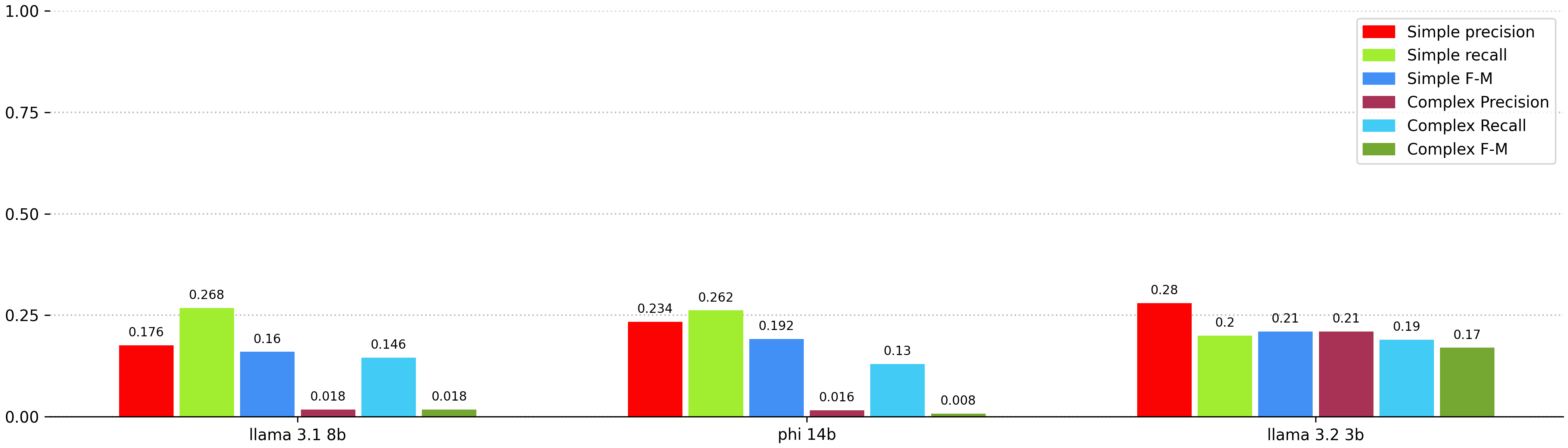}
    \caption{Performance of the models compared with the smaller but fine-tuned model.}
    \label{fig:c8:fine-prev}
\end{figure}

Another possibility of improvement is testing how fine-tuning can improve the LLM capacity of complex alignments and if it can generalize. To test this, the first experiment was to fine-tune the model Meta/LLaMA-3.2-3B-Instruct (3B) in the modules dataset with different proportions of train and test, and compare with the LLMs Meta/LLaMA-3.1-8B-Instruct (8B) and microsoft/phi-4 (14B) without training. The test was performed with 90\% train with a total reference size of 382 EDOAL files, 798 correspondences. With this split, 343 pairs are in train and 39 are in test. The model was trained for 100 epochs. The results of this test are present in Figure \ref{fig:c8:fine-prev}. It is possible to see that the fine-tuning improved the performance of the model, leading it to have better results in both simple and complex cases, even with the model being smaller. Those results show the potential of fine-tuning in improving the capacity of the models given high-quality training data. While having improvements, the performance is still low (below 0.5). One of the hypotheses of this limited improvement lies in the low amount of specific alignment data for training. To investigate this path, a synthetic data generation strategy described in the next section was performed.

\subsection{Dataset Generation and Performance across Prompting Strategies}

This section presents the results of the synthetic dataset generation process, followed by an analysis of the LLM's performance. The dataset generation process yielded a total of 6,650 alignment pairs: 4,650 pairs contained one or more correspondences, while the remaining 2,000 were empty alignments with no correspondences.

Parsing the raw textual output from the LLM revealed that a subset of the generated files contained syntactical errors. These issues primarily included:
\begin{itemize}
    \item Missing prefix declarations
    \item Missing ontology tags
    \item Entities without a prefix
    \item Invalid literal 
    \item EOS tokens
\end{itemize}

To address these issues, an automated script has been implemented to repair such errors by adding missing prefix declarations and correcting ontology tags. After applying these corrections, 4,407 of the 4,650 alignments (95\%) were successfully validated, leaving 243 (5\%) as invalid. Similarly, 1,892 of the 2,000 empty alignment pairs (95\%) were rendered valid. Cumulatively, the combined generation and repair pipeline demonstrated a high degree of reliability, producing syntactically valid data in 95\% of all cases.

\begin{table}[!ht]

\begin{tabular}{|lllllll|}
\hline
matcher & s-p  & s-r  & s-f  & c-p  & c-r  & c-f  \\
\hline
\multicolumn{7}{|c|}{Conference} \\
\hline
AMLC    & 0.00 & 0.01 & 0.00 & 0.13 & 0.20 & 0.15 \\
AROA    & 0.00 & 0.00 & 0.00 & 0.00 & 0.00 & 0.00 \\
CANARD  & 0.00 & 0.00 & 0.00 & 0.00 & 0.00 & 0.00 \\
b0 (Llama-3-1-8B)     & 0.03 & 0.19 & 0.05 & 0.06 & 0.10 & 0.05 \\
b0 (Llama-3-2-3B)   & 0.25 & 0.12 & 0.14 & 0.09 & 0.03 & 0.03 \\
b0 (Phi-4-mini)     & 0.00 & 0.00 & 0.00 & 0.00 & 0.00 & 0.00 \\
b0 (Qwen3-14B)      & 0.53 & \textbf{0.66} & \textbf{0.55} & \textbf{0.24} & \textbf{0.37} & \textbf{0.24} \\
b0 (Mistral-7B)     & 0.14 & 0.12 & 0.11 & 0.08 & 0.05 & 0.05 \\
b0 (phi-4)          & \textbf{0.62} & 0.47 & 0.48 & 0.23 & 0.19 & 0.13 \\
b1      & 0.05 & 0.11 & 0.07 & 0.02 & 0.03 & 0.02 \\
b2      & 0.50 & 0.25 & 0.33 & 0.00 & 0.15 & 0.00 \\
\hline
\end{tabular}
    \caption{Performance comparison of the proposed approach against other matchers on the Conference dataset. The table displays precision (p), recall (r), and F-measure (f) for both simple (s-p, s-r, s-f) and complex (c-p, c-r, c-f) correspondences.}
    \label{tab:c8:res-conf}

\end{table}

\begin{table}[!ht]

\begin{tabular}{|lllllll|}
\hline
matcher & s-p  & s-r  & s-f  & c-p  & c-r  & c-f  \\
\hline
\multicolumn{7}{|c|}{Enslaved} \\
\hline
AMLC    & 0.45 & \textbf{0.93} & \textbf{0.60} & \textbf{0.32} & 0.06 & 0.11 \\
AROA    & 0.00 & 0.00 & 0.00 & 0.00 & 0.00 & 0.00 \\
CANARD  & \textbf{0.54} & 0.68 & \textbf{0.60} & 0.18 & 0.06 & 0.09 \\
b0 (Llama-3-1-8B)      & 0.04 & 0.18 & 0.07 & 0.00 & 0.05 & 0.00 \\
b0 (Llama-3-2-3B)   & 0.00 & 0.00 & 0.00 & 0.00 & 0.00 & 0.00 \\
b0 (Phi-4-mini)     & 0.00 & 0.00 & 0.00 & 0.00 & 0.00 & 0.00 \\
b0 (Qwen3-14B)      & 0.12 & 0.32 & 0.18 & 0.21 & \textbf{0.08} & \textbf{0.12} \\
b0 (Mistral-7B)     & 0.00 & 0.00 & 0.00 & 0.00 & 0.00 & 0.00 \\
b0 (phi-4)          & 0.00 & 0.00 & 0.00 & 0.00 & 0.00 & 0.00 \\
b1      & 0.13 & 0.04 & 0.06 & 0.09 & 0.00 & 0.01 \\
b2      & 0.09 & 0.18 & 0.12 & 0.00 & 0.01 & 0.00 \\
\hline
\end{tabular}
    \caption{Performance comparison of the proposed approach against other matchers on the Enslaved dataset. The table displays precision (p), recall (r), and F-measure (f) for both simple (s-p, s-r, s-f) and complex (c-p, c-r, c-f) correspondences.}
    \label{tab:c8:res-ensv}

\end{table}

\begin{table}[!ht]

\begin{tabular}{|lllllll|}
\hline
matcher & s-p  & s-r  & s-f  & c-p  & c-r  & c-f  \\
\hline
\multicolumn{7}{|c|}{Geolink} \\
\hline
AMLC    & 0.00 & 0.00 & 0.00 & 0.07 & 0.00 & 0.01 \\
AROA    & \textbf{0.94} & \textbf{0.82} & \textbf{0.87} & \textbf{0.56} & 0.15 & 0.24 \\
CANARD  & 0.00 & 0.00 & 0.00 & 0.00 & 0.00 & 0.00 \\
b0 (Llama-3-1-8B)      & 0.02 & 0.05 & 0.03 & 0.00 & 0.02 & 0.00 \\
b0 (Llama-3-2-3B)   & 0.00 & 0.00 & 0.00 & 0.00 & 0.00 & 0.00 \\
b0 (Phi-4-mini)     & 0.04 & 0.00 & 0.00 & 0.00 & 0.00 & 0.00 \\
b0 (Qwen3-14B)      & 0.18 & 0.79 & 0.30 & 0.26 & \textbf{0.26} & \textbf{0.26} \\
b0 (Mistral-7B)     & 0.16 & 0.59 & 0.25 & 0.18 & 0.05 & 0.07 \\
b0 (phi-4)          & 0.27 & 0.14 & 0.18 & 0.29 & 0.02 & 0.04 \\
b1      & 0.51 & 0.16 & 0.24 & 0.08 & 0.01 & 0.01 \\
b2      & 0.51 & 0.61 & 0.55 & 0.00 & 0.02 & 0.00 \\

\hline
\end{tabular}
    \caption{Performance comparison of the proposed approach against other matchers on the Geolink dataset. The table displays precision (p), recall (r), and F-measure (f) for both simple (s-p, s-r, s-f) and complex (c-p, c-r, c-f) correspondences.}
    \label{tab:c8:res-geo}

\end{table}

\begin{table}[!ht]

\begin{tabular}{|lllllll|}
\hline
matcher & s-p  & s-r  & s-f  & c-p  & c-r  & c-f  \\
\hline
\multicolumn{7}{|c|}{Hydrography} \\
\hline
AMLC    & 0.02 & 0.00 & 0.00 & 0.02 & 0.02 & 0.02 \\
AROA    & 0.00 & 0.00 & 0.00 & 0.00 & 0.00 & 0.00 \\
CANARD  & 0.00 & 0.00 & 0.00 & 0.00 & 0.00 & 0.00 \\
b0 (Llama-3-1-8B)      & 0.30 & 0.13 & 0.15 & 0.08 & 0.14 & 0.06 \\
b0 (Llama-3-2-3B)   & 0.04 & 0.00 & 0.01 & 0.00 & 0.00 & 0.00 \\
b0 (Phi-4-mini)     & 0.00 & 0.00 & 0.00 & 0.00 & 0.00 & 0.00 \\
b0 (Qwen3-14B)      & \textbf{0.41} & \textbf{0.46} & \textbf{0.41} & 0.39 & \textbf{0.24} & \textbf{0.28} \\
b0 (Mistral-7B)     & \textbf{0.41} & 0.10 & 0.13 & 0.18 & 0.11 & 0.04 \\
b0 (phi-4)          & 0.36 & 0.35 & 0.32 & 0.27 & \textbf{0.24} & 0.16 \\
b1      & 0.27 & 0.04 & 0.07 & \textbf{0.43} & 0.04 & 0.07 \\
b2      & 0.37 & 0.18 & 0.24 & 0.00 & 0.16 & 0.00 \\
\hline
\end{tabular}
    \caption{Performance comparison of the proposed approach against other matchers on the Hydrography dataset. The table displays precision (p), recall (r), and F-measure (f) for both simple (s-p, s-r, s-f) and complex (c-p, c-r, c-f) correspondences.}
    \label{tab:c8:res-hydro}

\end{table}

\begin{table}[!ht]

\begin{tabular}{|lllllll|}
\hline
matcher & s-p  & s-r  & s-f  & c-p  & c-r  & c-f  \\
\hline
\multicolumn{7}{|c|}{Taxon} \\
\hline
AMLC    & 0.00 & 0.00 & 0.00 & 0.00 & 0.00 & 0.00 \\
AROA    & 0.00 & 0.00 & 0.00 & 0.00 & 0.00 & 0.00 \\
CANARD  & \textbf{0.35} & 0.02 & 0.03 & \textbf{0.38} & \textbf{0.34} & \textbf{0.34} \\
b0 (Llama-3-1-8B)      & 0.04 & 0.02 & 0.01 & 0.26 & 0.10 & 0.14 \\
b0 (Llama-3-2-3B)   & 0.00 & 0.00 & 0.00 & 0.00 & 0.00 & 0.00 \\
b0 (Phi-4-mini)     & 0.00 & 0.00 & 0.00 & 0.00 & 0.00 & 0.00 \\
b0 (Qwen3-14B)      & 0.19 & \textbf{0.09} & \textbf{0.12} & 0.05 & 0.17 & 0.07 \\
b0 (Mistral-7B)     & 0.05 & 0.00 & 0.00 & 0.00 & 0.02 & 0.00 \\
b0 (phi-4)          & 0.21 & 0.00 & 0.00 & 0.00 & 0.14 & 0.00 \\
b1      & 0.06 & 0.00 & 0.00 & 0.27 & 0.03 & 0.06 \\
b2      & 0.19 & 0.02 & 0.03 & 0.00 & 0.02 & 0.00 \\

\hline
\end{tabular}
    \caption{Performance comparison of the proposed approach against other matchers on the Taxon dataset. The table displays precision (p), recall (r), and F-measure (f) for both simple (s-p, s-r, s-f) and complex (c-p, c-r, c-f) correspondences.}
    \label{tab:c8:res-taxon}

\end{table}

The full evaluation results are presented in Table \ref{tab:c8:res-conf} for Conference, Table \ref{tab:c8:res-ensv} for Enslaved, Table \ref{tab:c8:res-geo} for GeoLink, Table \ref{tab:c8:res-hydro} for Hydrography, and Table \ref{tab:c8:res-taxon} for Taxon, where a score of zero indicates that a matcher failed to produce any alignments for a given dataset. A breakdown of the top performers shows that in the Conference dataset, the proposed setting variant b0 with Qwen3-14B led in simple matching with an F-measure of 0.55 and also in complex matching with a F-measure of 0.24, but the variant b0 with phi-4 has the highest precision in simple matching with 0.62. For the Enslaved dataset, AMLC \cite{DBLP:conf/semweb/LimaFCCP20} and CANARD tied for the lead in simple matching with an F-measure of 0.60, and AMLC also topped the complex alignments precision with 0.32, and the top complex F-measure is the variant b0 Qwen-14B with 0.12. AROA \cite{DBLP:conf/semweb/ZhouH20} dominated the Geolink dataset in simple with 0.87 f-measure and highest precision in complex with 0.56, but in complex f-measure, the b0 with Qwen3-14B has the highest value with 0.26. In the Hydrography dataset, the b0 variation with Qwen3-14B dominates in both simple and complex f-measures with 0.41 and 0.28, respectively. Finally, the b0 variant with Qwen3-14B has the highest f-measure in the simple case with 0.09, but in the complex case, CANARD has the highest results with 0.34 f-measure. These results reveal several trends. First, different systems tend to excel on different datasets, showing a high degree of matcher specialization with no single best performer across all datasets. Second, as expected, performance on complex matching is still lower than on simple matching for all systems. Finally, generalization remains a significant challenge, as many matchers do not produce results for all datasets. A clear example is CANARD, which requires instances to run and therefore fails on the instance-free Conference and Geolink datasets, resulting in a zero score.

\begin{figure}
    \centering
    \includegraphics[width=\linewidth]{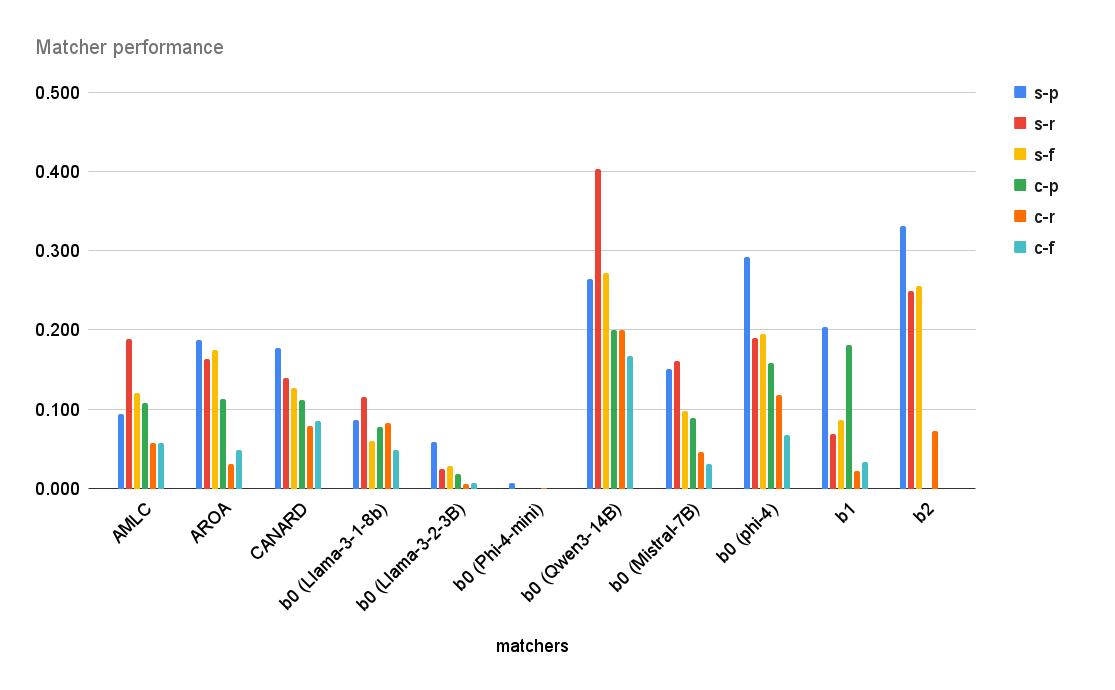}
    \caption{Average performance of the proposed approaches across all evaluated datasets. The table shows precision (p), recall (r), and F-measure (f) for both simple (s-p, s-r, s-f) and complex (c-p, c-r, c-f) correspondences.}
    \label{fig:gen}
\end{figure}

To assess the generalization capabilities of the matchers, their average performance across all datasets is presented in Figure \ref{fig:gen}. On average, the proposed method with the b0 variation with Qwen3-14B achieved the highest scores for the simple alignments and for the complex alignments of the complex track. The results confirm that the LLM is capable of doing direct matching if the right modularization is found. Moreover, fine-tuning provides a performance boost over the base model for these types of correspondences. However, this approach, in the way it was implemented, comes with clear trade-offs. The b2 model, for example, was limited to generating only 1:1 correspondences, which decreased its performance on complex matching tasks. In contrast, the cross-validation variant successfully increased the model's precision, but this gain came at the cost of a lower recall.

\subsection{Discussion and Implications}

These results confirm the significant potential of leveraging LLMs for ontology matching. With proper training and carefully designed prompts, these models can achieve strong generalization and performance, even when applied in a zero-shot setting. However, key challenges remain. The limited recall and consistently low performance on complex n:m correspondences indicate that further research is required to produce the high-quality, robust alignments needed for real-world scenarios.

The approach proposed in this paper, instruction fine-tuning on domain-specific data, is a direct step toward bridging this performance gap. By providing a robust method for generating subontology training pairs, the proposed method can be configured to address specific challenges. In particular, this adaptability opens future avenues for creating training data that explicitly targets n:m alignments and even multilingual contexts, paving the way for more powerful and versatile LLM-based ontology matchers.

\section{Conclusion}
\label{conclusion}

These results confirm the significant potential of leveraging LLMs for ontology matching. It is possible to see that the size of the LLM impacts the matching performance, also with the usage of reasoning models. With proper training and carefully designed prompts, these models can achieve strong generalization and performance, even when applied in a zero-shot setting. However, key challenges remain. The limited recall and consistently low performance on fine-tuning for complex alignments indicate that further research is required to produce the high-quality, robust alignments needed for real-world scenarios. In addition, improving the quality of the space reduction module is required to achieve the highest results as found in this experiment.

The evaluation across five datasets from the OAEI Complex Track demonstrated that fine-tuning on the generated data (b2) significantly improved performance for simple 1:1 alignments, while cross-validation fine-tuning (b1) increased precision for complex alignments. These results highlight that synthetic training data can enhance the adaptability of LLMs for ontology matching, even when manually aligned examples from the target domain are unavailable.

Nevertheless, the experiments also revealed that performance on complex alignments remains limited, especially for n:m correspondences. While the fine-tuned models achieved higher precision and recall on some datasets, generalization across all benchmarks remains a significant challenge. This suggests that further advances are needed in two key areas: the generation of training data that better captures complex alignment structures and the design of prompts and model architectures tailored for structured semantic reasoning.

Future work will focus on several key directions. It is possible to expand the synthetic data generation process to include the verification of the integrity and semantic equivalence of correspondences.  It is also possible to extend its coverage to a broader variety of logical constructors and multilingual scenarios. Furthermore, more improvements can come from integrating explicit reasoning mechanisms into the LLM pipeline and exploring hybrid approaches that combine LLM-based generation with traditional ontology matching techniques. Through these efforts, it is expected to bridge the current performance gap in complex alignment quality and advance towards more robust, scalable, and generalizable ontology matching solutions powered by LLMs.

\printbibliography

@inproceedings{DBLP:conf/semweb/ZhouH20,
  author       = {Lu Zhou and
                  Pascal Hitzler},
  editor       = {Pavel Shvaiko and
                  J{\'{e}}r{\^{o}}me Euzenat and
                  Ernesto Jim{\'{e}}nez{-}Ruiz and
                  Oktie Hassanzadeh and
                  C{\'{a}}ssia Trojahn},
  title        = {{AROA} results for {OAEI} 2020},
  booktitle    = {Proceedings of the 15th International Workshop on Ontology Matching
                  co-located with the 19th International Semantic Web Conference {(ISWC}
                  2020), Virtual conference (originally planned to be in Athens, Greece),
                  November 2, 2020},
  series       = {{CEUR} Workshop Proceedings},
  volume       = {2788},
  pages        = {161--167},
  publisher    = {CEUR-WS.org},
  year         = {2020},
  url          = {https://ceur-ws.org/Vol-2788/oaei20\_paper4.pdf},
  bibsource    = {dblp computer science bibliography, https://dblp.org}
}

@inproceedings{DBLP:conf/semweb/LimaFCCP20,
  author       = {Beatriz Lima and
                  Daniel Faria and
                  Francisco M. Couto and
                  Isabel F. Cruz and
                  Catia Pesquita},
  editor       = {Pavel Shvaiko and
                  J{\'{e}}r{\^{o}}me Euzenat and
                  Ernesto Jim{\'{e}}nez{-}Ruiz and
                  Oktie Hassanzadeh and
                  C{\'{a}}ssia Trojahn},
  title        = {{OAEI} 2020 results for {AML} and {AMLC}},
  booktitle    = {Proceedings of the 15th International Workshop on Ontology Matching
                  co-located with the 19th International Semantic Web Conference {(ISWC}
                  2020), Virtual conference (originally planned to be in Athens, Greece),
                  November 2, 2020},
  series       = {{CEUR} Workshop Proceedings},
  volume       = {2788},
  pages        = {154--160},
  publisher    = {CEUR-WS.org},
  year         = {2020},
  url          = {https://ceur-ws.org/Vol-2788/oaei20\_paper3.pdf},
  bibsource    = {dblp computer science bibliography, https://dblp.org}
}

@inproceedings{DBLP:conf/esws/SousaLS25,
	title        = {On Evaluation Metrics for Complex Matching Based on Reference Alignments},
	author       = {Guilherme Henrique Santos Sousa and Rinaldo Lima and C{\'{a}}ssia Trojahn dos Santos},
	year         = 2025,
	booktitle    = {The Semantic Web - 22nd European Semantic Web Conference, {ESWC} 2025, Portoroz, Slovenia, June 1-5, 2025, Proceedings, Part {I}},
	publisher    = {Springer},
	series       = {Lecture Notes in Computer Science},
	volume       = 15718,
	pages        = {77--93},
	doi          = {10.1007/978-3-031-94575-5\_5},
	url          = {https:doi.org/10.1007/978-3-031-94575-5\_5},
	editor       = {Edward Curry and Maribel Acosta and Mar{\'{\i}}a Poveda{-}Villal{\'{o}}n and Marieke van Erp and Adegboyega K. Ojo and Katja Hose and Cogan Shimizu and Pasquale Lisena},
	bibsource    = {dblp computer science bibliography, https:dblp.org}
}

@phdthesis{scharffe2009correspondence,
  title={Correspondence patterns representation},
  author={Scharffe, Fran{\c{c}}ois},
  year={2009},
  school={PhD thesis, University of Innsbruck}
}

@inproceedings{DBLP:conf/kgswc/AminiNHA24,
  author       = {Reihaneh Amini and
                  Sanaz Saki Norouzi and
                  Pascal Hitzler and
                  Reza Amini},
  editor       = {Sanju Tiwari and
                  Boris Villaz{\'{o}}n{-}Terrazas and
                  Fernando Ortiz{-}Rodr{\'{\i}}guez and
                  Soror Sahri},
  title        = {Towards Complex Ontology Alignment Using Large Language Models},
  booktitle    = {Knowledge Graphs and Semantic Web - 6th International Conference,
                  {KGSWC} 2024, Paris, France, December 11-13, 2024, Proceedings},
  series       = {Lecture Notes in Computer Science},
  volume       = {15459},
  pages        = {17--31},
  publisher    = {Springer},
  year         = {2024},
  url          = {https://doi.org/10.1007/978-3-031-81221-7\_2},
  doi          = {10.1007/978-3-031-81221-7\_2},
  bibsource    = {dblp computer science bibliography, https://dblp.org}
}

@inproceedings{DBLP:conf/om2/SousaLT24,
  author       = {Guilherme Henrique Santos Sousa and
                  Rinaldo Lima and
                  C{\'{a}}ssia Trojahn},
  editor       = {Ernesto Jim{\'{e}}nez{-}Ruiz and
                  Oktie Hassanzadeh and
                  C{\'{a}}ssia Trojahn and
                  Sven Hertling and
                  Huanyu Li and
                  Pavel Shvaiko and
                  J{\'{e}}r{\^{o}}me Euzenat},
  title        = {Towards Generating Complex Alignments with Large Language Models via
                  Prompt Engineering},
  booktitle    = {Proceedings of the 19th International Workshop on Ontology Matching
                  co-located with the 23rd International Semantic Web Conference {(ISWC}
                  2024), Baltimore, USA, November 11, 2024},
  series       = {{CEUR} Workshop Proceedings},
  volume       = {3897},
  pages        = {43--56},
  publisher    = {CEUR-WS.org},
  year         = {2024},
  url          = {https://ceur-ws.org/Vol-3897/om2024\_LTpaper4.pdf},
  bibsource    = {dblp computer science bibliography, https://dblp.org}
}

@inproceedings{DBLP:conf/iclr/WeiBZGYLDDL22,
  author       = {Jason Wei and
                  Maarten Bosma and
                  Vincent Y. Zhao and
                  Kelvin Guu and
                  Adams Wei Yu and
                  Brian Lester and
                  Nan Du and
                  Andrew M. Dai and
                  Quoc V. Le},
  title        = {Finetuned Language Models are Zero-Shot Learners},
  booktitle    = {The Tenth International Conference on Learning Representations, {ICLR}
                  2022, Virtual Event, April 25-29, 2022},
  publisher    = {OpenReview.net},
  year         = {2022},
  url          = {https://openreview.net/forum?id=gEZrGCozdqR},
  bibsource    = {dblp computer science bibliography, https://dblp.org}
}

@article{DBLP:journals/corr/abs-2502-13619,
  author       = {Guilherme Sousa and
                  Rinaldo Lima and
                  C{\'{a}}ssia Trojahn},
  title        = {Complex Ontology Matching with Large Language Model Embeddings},
  journal      = {CoRR},
  volume       = {abs/2502.13619},
  year         = {2025},
  url          = {https://doi.org/10.48550/arXiv.2502.13619},
  doi          = {10.48550/ARXIV.2502.13619},
  eprinttype    = {arXiv},
  eprint       = {2502.13619},
  bibsource    = {dblp computer science bibliography, https://dblp.org}
}

@article{DBLP:journals/semweb/ThieblinSHT24,
  author       = {{\'{E}}lodie Thi{\'{e}}blin and
                  Guilherme Sousa and
                  Ollivier Haemmerl{\'{e}} and
                  C{\'{a}}ssia Trojahn},
  title        = {{CANARD:} An approach for generating expressive correspondences based
                  on competency questions for alignment},
  journal      = {Semantic Web},
  volume       = {15},
  number       = {3},
  pages        = {897--929},
  year         = {2024},
  url          = {https://doi.org/10.3233/SW-233521},
  doi          = {10.3233/SW-233521},
  bibsource    = {dblp computer science bibliography, https://dblp.org}
}

@inproceedings{DBLP:conf/ecai/SilvaFP24,
  author       = {Marta Contreiras Silva and
                  Daniel Faria and
                  Catia Pesquita},
  editor       = {Ulle Endriss and
                  Francisco S. Melo and
                  Kerstin Bach and
                  Alberto Jos{\'{e}} Bugar{\'{\i}}n Diz and
                  Jose Maria Alonso{-}Moral and
                  Sen{\'{e}}n Barro and
                  Fredrik Heintz},
  title        = {Complex Multi-Ontology Alignment Through Geometric Operations on Language
                  Embeddings},
  booktitle    = {{ECAI} 2024 - 27th European Conference on Artificial Intelligence,
                  19-24 October 2024, Santiago de Compostela, Spain - Including 13th
                  Conference on Prestigious Applications of Intelligent Systems {(PAIS}
                  2024)},
  series       = {Frontiers in Artificial Intelligence and Applications},
  volume       = {392},
  pages        = {1333--1340},
  publisher    = {{IOS} Press},
  year         = {2024},
  url          = {https://doi.org/10.3233/FAIA240632},
  doi          = {10.3233/FAIA240632},
  bibsource    = {dblp computer science bibliography, https://dblp.org}
}

@article{DBLP:journals/semweb/DavidESS11,
  author       = {J{\'{e}}r{\^{o}}me David and
                  J{\'{e}}r{\^{o}}me Euzenat and
                  Fran{\c{c}}ois Scharffe and
                  C{\'{a}}ssia Trojahn dos Santos},
  title        = {The Alignment {API} 4.0},
  journal      = {Semantic Web},
  volume       = {2},
  number       = {1},
  pages        = {3--10},
  year         = {2011},
  url          = {https://doi.org/10.3233/SW-2011-0028},
  doi          = {10.3233/SW-2011-0028},
  bibsource    = {dblp computer science bibliography, https://dblp.org}
}

@inproceedings{DBLP:conf/kcap/HertlingP23,
  author       = {Sven Hertling and
                  Heiko Paulheim},
  editor       = {Kristen Brent Venable and
                  Daniel Garijo and
                  Brian Jalaian},
  title        = {OLaLa: Ontology Matching with Large Language Models},
  booktitle    = {Proceedings of the 12th Knowledge Capture Conference 2023, {K-CAP}
                  2023, Pensacola, FL, USA, December 5-7, 2023},
  pages        = {131--139},
  publisher    = {{ACM}},
  year         = {2023},
  url          = {https://doi.org/10.1145/3587259.3627571},
  doi          = {10.1145/3587259.3627571},
  bibsource    = {dblp computer science bibliography, https://dblp.org}
}

@inproceedings{DBLP:conf/esws/GiglouDEA24,
  author       = {Hamed Babaei Giglou and
                  Jennifer D'Souza and
                  Felix Engel and
                  S{\"{o}}ren Auer},
  editor       = {Albert Mero{\~{n}}o{-}Pe{\~{n}}uela and
                  {\'{O}}scar Corcho and
                  Paul Groth and
                  Elena Simperl and
                  Valentina Tamma and
                  Andrea Giovanni Nuzzolese and
                  Mar{\'{\i}}a Poveda{-}Villal{\'{o}}n and
                  Marta Sabou and
                  Valentina Presutti and
                  Irene Celino and
                  Artem Revenko and
                  Joe Raad and
                  Bruno Sartini and
                  Pasquale Lisena},
  title        = {LLMs4OM: Matching Ontologies with Large Language Models},
  booktitle    = {The Semantic Web: {ESWC} 2024 Satellite Events - Hersonissos, Crete,
                  Greece, May 26-30, 2024, Proceedings, Part {I}},
  series       = {Lecture Notes in Computer Science},
  volume       = {15344},
  pages        = {25--35},
  publisher    = {Springer},
  year         = {2024},
  url          = {https://doi.org/10.1007/978-3-031-78952-6\_3},
  doi          = {10.1007/978-3-031-78952-6\_3},
  bibsource    = {dblp computer science bibliography, https://dblp.org}
}

@inproceedings{DBLP:conf/esws/GiglouDKA25,
  author       = {Hamed Babaei Giglou and
                  Jennifer D'Souza and
                  Oliver Karras and
                  S{\"{o}}ren Auer},
  editor       = {Edward Curry and
                  Maribel Acosta and
                  Mar{\'{\i}}a Poveda{-}Villal{\'{o}}n and
                  Marieke van Erp and
                  Adegboyega K. Ojo and
                  Katja Hose and
                  Cogan Shimizu and
                  Pasquale Lisena},
  title        = {OntoAligner: {A} Comprehensive Modular and Robust Python Toolkit for
                  Ontology Alignment},
  booktitle    = {The Semantic Web - 22nd European Semantic Web Conference, {ESWC} 2025,
                  Portoroz, Slovenia, June 1-5, 2025, Proceedings, Part {II}},
  series       = {Lecture Notes in Computer Science},
  volume       = {15719},
  pages        = {174--191},
  publisher    = {Springer},
  year         = {2025},
  url          = {https://doi.org/10.1007/978-3-031-94578-6\_10},
  doi          = {10.1007/978-3-031-94578-6\_10},
  bibsource    = {dblp computer science bibliography, https://dblp.org}
}

@Book{jm3,
  author =       "Daniel Jurafsky and James H. Martin",
  title =        "Speech and Language Processing: An Introduction to
                 Natural Language Processing, Computational Linguistics,
                 and Speech Recognition with Language Models",
  year =         "2025",
  url = {https://web.stanford.edu/~jurafsky/slp3/},
  note = "Online manuscript released January 12, 2025",
  edition =         "3rd",
}

\end{document}